\documentclass[10pt,twocolumn,letterpaper]{article}

\usepackage{cvpr}
\usepackage{times}
\usepackage{epsfig}
\usepackage{graphicx}
\usepackage{amsmath}
\usepackage{amssymb}

\usepackage{multirow}
\usepackage{adjustbox}
\usepackage{tabu}
\usepackage{tabularx}
\usepackage{xcolor}
\usepackage{soul}
\usepackage[linesnumbered, ruled]{algorithm2e}
\usepackage{algpseudocode}
\usepackage{bm}
\usepackage{array}
\usepackage{enumitem}
\usepackage{verbatim}
\usepackage{subfigure}
\usepackage{caption}
\usepackage{algorithmicx}
\usepackage{comment}
\usepackage{lipsum}
\usepackage{stringstrings}

\def\bm{{\bf m}}
\def\bbf{{\mathbf{f}}}
\def\mbR{{\mathbb R}}

\def\cK{{\mathcal K}}

\def\bMI{{\mathbf{M}_\text{I} }}
\def\bMD{{\mathbf{M}_\text{D} }}
\def\tI{ {\text{I}} }
\def\sp{{ ~ }}

\usepackage[margin=4pt,font=footnotesize,labelfont=bf,labelsep=endash,tableposition=top]{caption}
\usepackage[pagebackref=true,breaklinks=true,letterpaper=true,colorlinks,bookmarks=false]{hyperref}
\cvprfinalcopy

\begin{document}

\title{Memorizing Comprehensively to Learn Adaptively: Unsupervised\\ Cross-Domain Person Re-ID with Multi-level Memory}

\author{
	Xinyu Zhang \textsuperscript{1,2}$^{*}$        \sp
	Dong Gong \textsuperscript{2}\thanks{The first two authors contribute equally. This work was done when X. Zhang was visiting The University of Adelaide.}       \sp
	Jiewei Cao \textsuperscript{2}       \sp
	Chunhua Shen \textsuperscript{2}\thanks{Corresponding author: $\tt chunhua.shen@adelaide.edu.au $.}   \sp
	\\
	$^{1}~$Tongji University, China 
	~
    $^{2}~$The University of Adelaide, Australia
}

\maketitle

\begin{abstract}
Unsupervised cross-domain person re-identification (Re-ID) aims to adapt the information from the labelled source domain to an unlabelled target domain. Due to the lack of supervision in the target domain, it is crucial to identify the underlying similarity-and-dissimilarity relationships among the unlabelled samples in the target domain. In order to use the whole data relationships efficiently in mini-batch training, we apply a series of memory modules to maintain an up-to-date representation of the entire dataset. Unlike the simple exemplar memory in previous works, we propose a novel multi-level memory network (MMN) to discover multi-level complementary information in the target domain, relying on three memory modules, i.e., part-level memory, instance-level memory, and domain-level memory. The proposed memory modules store multi-level representations of the target domain, which capture both the fine-grained differences between images and the global structure for the holistic target domain. The three memory modules complement each other and systematically integrate multi-level supervision from bottom to up. Experiments on three datasets demonstrate that the multi-level memory modules cooperatively boost the unsupervised cross-domain Re-ID task, and the proposed MMN achieves competitive results.
\end{abstract}

\begin{figure}[!t]
\centering
\includegraphics[trim =0mm 0mm 0mm 0mm, clip, width=.95\linewidth]{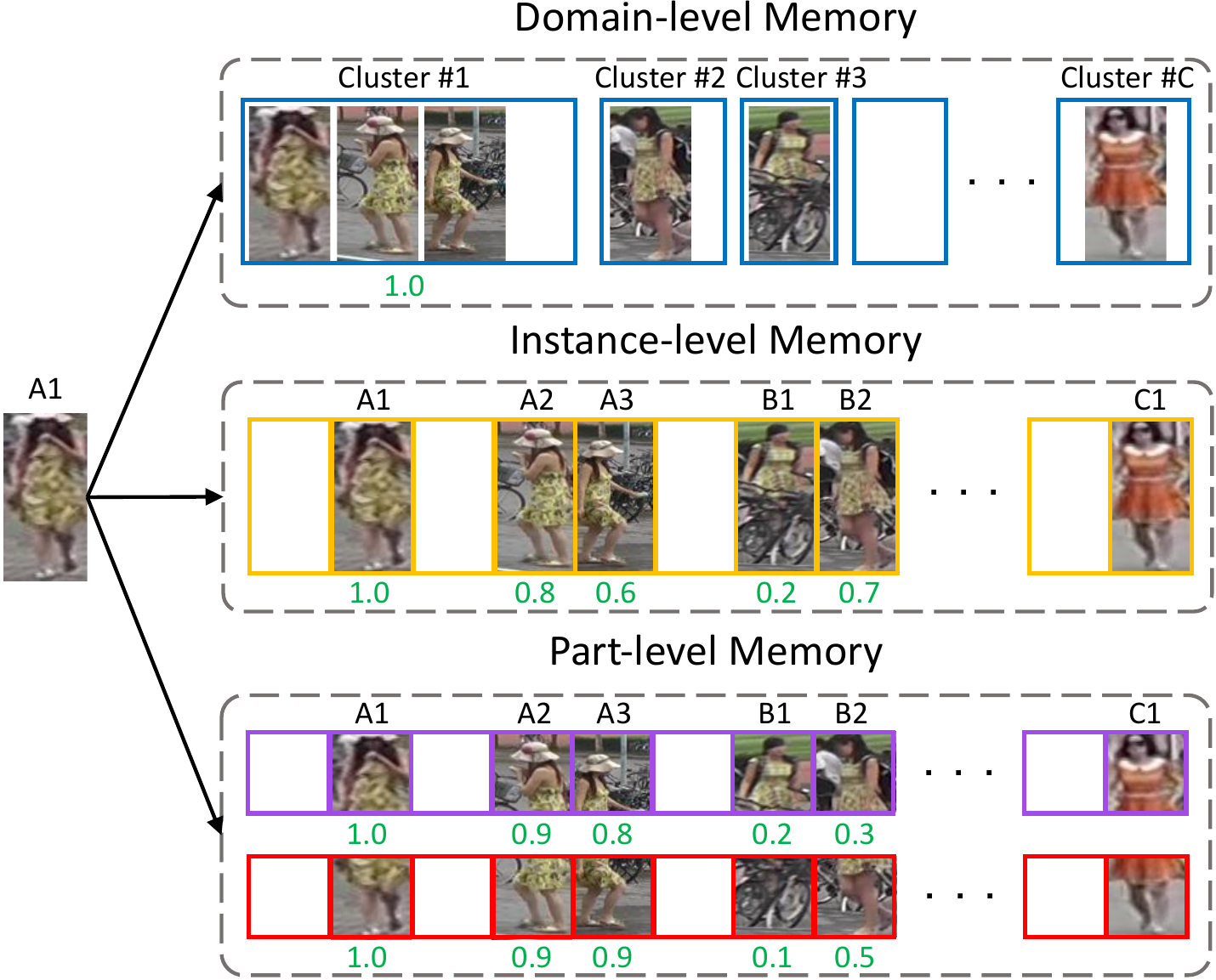}
\setlength{\abovecaptionskip}{0.2cm} 
\setlength{\belowcaptionskip}{-0.3cm}
\caption{
Multi-level memories cooperatively maintain a full representation of the target domain.
A, B and C denote different identities in a real dataset. 
The \textcolor[rgb]{0.2,0.7,0.3}{green} values represent the similarity between A1 and other samples. 
In the confusing scenario, only using instance-level memory mistakenly produces high similarity between A1 and B2.
However, the dissimilarity between A1 and B2 can be reflected by focusing on the upper body. 
At the other hand, A1 and B2 belong to different clusters in the domain-level memory.
It shows that MMN can discover more reliable similarity-and-dissimilarity relationships from different-level memory modules. 
The proposed memory model is crucial as a hub for integrating the information from multiple levels.
}
\label{fig:figure1}
\vspace{-3.1mm}
\end{figure}

\section{Introduction}
Person re-identification (Re-ID) is a crucial task aiming to match a specific person image with other images of this person across non-overlapping camera views. 
Most of the existing methods draw much
attention on the feature representation~\cite{chang2018multi,zheng2016personsurvery,sun2018beyond,Li_2018_CVPR,chang2018multi,xu2018attention,zhang2017alignedreid,chen2017person,sun2017svdnet,chenself,chen2019mixed} and metric learning~\cite{cheng2016person,hermans2017defense,chen2017beyond} under the \textit{supervised learning} setting. 
Although these methods achieve impressive performances, they highly depend on abundant labelled data which can be costly and time-consuming to obtain.

Some recent approaches~\cite{huang2018eanet,deng2018image,fan2017pul,peng2016unsupervised,wang2018reid,wei2018person,Zhong_2018_ECCV} attempt to study \textit{unsupervised cross-domain} person Re-ID, which mainly focuses on how to adapt the information from the labelled \textit{source domain} to an unlabelled \textit{target domain} so as to take advantage of large and easily collected unlabelled data.
The main difficulties are two-fold. 
Firstly, data points from different domains suffer from \textit{domain shift} \cite{zhang2019self,song2019generalizable,wei2018person,Zhong_2018_ECCV}, \ie, inconsistent data distributions. 
Some methods \cite{deng2018image,wei2018person,bak2018domain,li2018adaptation} 
spend effort
to narrow down the domain gaps 
using 
Generative Adversarial Network (GAN).
The other problem is \textit{intra-domain variation}, which is more crucial for the performance due to lacking identity annotation on the target data. 
To handle this issue, some methods try to discover the underlying relationships among the unlabelled data points \cite{zhong2019invariance,fu2019self,zhang2019self}. 
To use the invariance properties over the whole training set in the mini-batch training, ECN \cite{zhong2019invariance} maintain an exemplar memory to record features of the target samples.
It identifies the sample relationships by solely measuring the exemplar-level similarity, which 
is often
fragile and can be misled easily, since it ignores the \textit{fine-grained} information between similar but different identities as well as the \textit{global distribution} of the data points with large variances in the same identities.
Although SSG~\cite{fu2019self} and PAST~\cite{zhang2019self} apply part-based clustering methods to capture the global and local information, they mainly rely on some heuristic training techniques and do not fully capture the global data distribution.

\par
In order to tackle above issues, we propose a novel \textit{multi-level memory network} (MMN) to use multi-level information of the target domain.
Memory modules are used to maintain representations of the entire dataset, which are updated dynamically and enable the model to use the whole data relationships in each mini-batch efficiently. 
Beyond the exemplar memory in \cite{zhong2019invariance}, MMN maintains memory pools on three representation levels, \ie, 
\textit{part-level memory},
\textit{instance-level memory},  and \textit{domain-level memory}. 
The three memory modules complement each other by integrating the information from different levels in a systematical scheme, instead of using them heuristically. 
It empowers MMN to discover the similarity-and-dissimilarity relationships among the target samples and use 
them to provide more reliable supervision for unlabelled samples. 

\par
The most straightforward approach 
to build memory on the target examples is to maintain the image-based representations in the slots as \cite{zhong2019invariance}.  
Each slot in our instance-level memory thus stores the entire image features that encodes the whole body information of a person. 
Given an input, the most similar memory items are selected (based on $k$-nearest neighbor search) as the ones with same identities to provide supervision signals. 
However, the similarity measuring may contain errors and mislead training, since the samples with similar overall appearance can belong to different identities (see Figure \ref{fig:figure1}) and the similarity only focusing on independent samples is fragile. We thus introduce \textit{part-level memory} and \textit{domain-level memory} to handle the issues, respectively. 
Considering that some subtle differences of different identifies can be distinguished by focusing on local parts \cite{su2017pose,suh2018part,sun2018beyond,zhao2017spindle,zhao2017deeply,zhang2019densely,sun2019perceive,yang2019patch,fu2019self}, in the part-level memory, each slot maintains the features of separate parts of a sample. 
We use the part-level memory to verify and rectify the similarity relationships obtained from the instance-level memory, instead of using it independently. 
The \textit{domain-level memory} is introduced to fully capture the overall structure of the target domain. 
Specifically, we cluster data points into several pseudo classes and assign each memory slot as the cluster centroid, \ie, the representative prototype of a cluster. 
The domain-level memory can further guide the 
memory read operation in instance level and part level and directly enforce the model to concentrate more on discriminative features on a high level. 

The three multi-level memory modules are trained under mutual guidance from each other and work cooperatively to improve the generalization ability of the model. 
To summarize, our main contributions are as follows:
\setlist{nolistsep}
\begin{itemize}[fullwidth, itemindent=1em]
\item[1)] We introduce a dynamically updated multi-level memory network (MMN) to systematically capture the multi-level information from bottom to up in the whole target domain.
\item[2)] The three memory modules in MMN are complementary to each other, which considers the whole data relationships from both fine-grained information and holistic structure simultaneously. 
Particularly, 
the part-level memory rectifies the similarity relationships for instance-level memory, while the domain-level memory provides the guidance of memory read for the other two. 
\item[3)] 
Experiments on three large-scale datasets demonstrate the effectiveness of our MMN on the task
of unsupervised cross-domain person Re-ID.
\end{itemize}
\setlist{nolistsep}

\begin{figure*}[t]
\centering
\includegraphics[trim =0mm 0mm 0mm 0mm, clip, width=.9\linewidth]{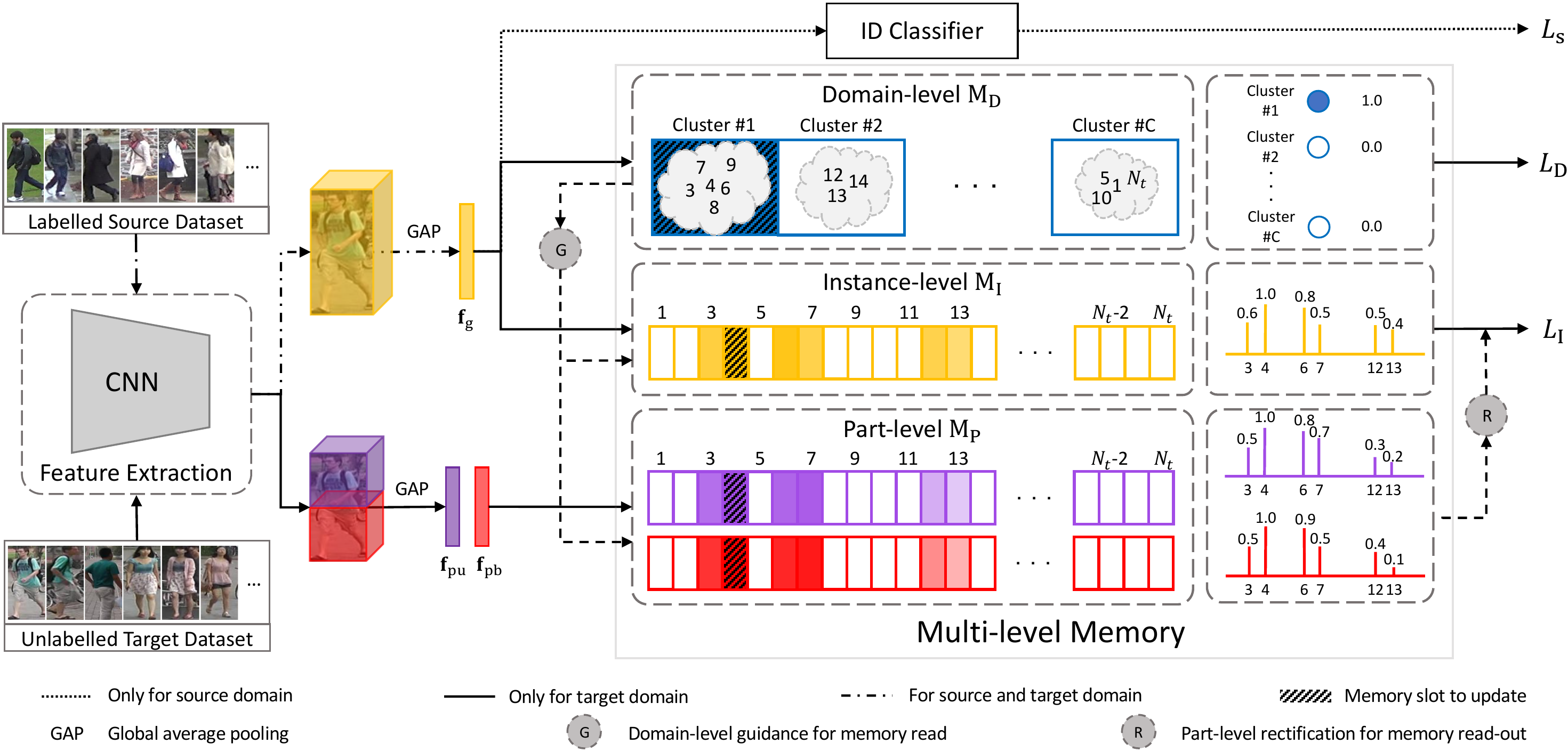} 
\setlength{\abovecaptionskip}{0.13cm} 
\setlength{\belowcaptionskip}{-0.3cm}
\caption{Diagram of our proposed multi-level memory network (MMN).
The source domain branch applies a classifier to calculate the softmax cross-entropy loss $L_{\text{s}}$.
The target domain branch is with a multi-level memory module, in which two loss functions $L_\text{I}$ and $L_\text{D}$ perform on the instance-level memory and domain-level memory, respectively.
Specifically, the three memories cooperatively provide reliable supervision for the target data from fine-grained information to holistic representation.
These memory modules are complementary with each other,
\ie, the domain-level guidance for memory read (Section~\ref{sec:domain}) and the part-level rectification for memory read-out (Section~\ref{sec:part}).}  
\label{fig:model}
\end{figure*}

\section{Related Work}
\noindent\textbf{Part-based Person Re-ID.} Most existing methods~\cite{sun2018beyond,su2017pose,zhao2017deeply,suh2018part,zhao2017spindle,bai2017scalable} attempt to take advantage of local structures to improve the holistic feature representation. Among these methods, \cite{sun2018beyond,wang2018learning} split the feature maps evenly into several parts from vertical and horizontal orientations.
Others like~\cite{sun2019perceive,su2017pose,zhao2017spindle,saquib2018pose,zheng2019pose,insafutdinov2016deepercut,kalayeh2018human,zhang2019densely} utilize various part-location algorithms to further advance the retrieval accuracy via extracting more accurate part regions.
For example, VPM~\cite{sun2019perceive} explores the visible regions through probability maps to locate the shared parts between images. 
\cite{su2017pose,zhao2017spindle,saquib2018pose,zheng2019pose,insafutdinov2016deepercut} learn to extract relatively precise part regions via pose estimation algorithms.
\cite{kalayeh2018human,zhang2019densely} learn to employ human semantic information to fully use the well-aligned local representation of the human body. 
Although these methods have achieved great improvement, they are all designed for supervised learning and cannot generalize well to the unlabelled datasets.

\noindent\textbf{Unsupervised Cross-Domain Person Re-ID.}
Recently, many methods~\cite{wei2018person,deng2018image,Zhong_2018_ECCV,wang2018reid,huang2018eanet,li2018unsupervised,peng2016unsupervised,liu2017stepwise,song2018unsupervised,zhang2019self,fu2019self} pay more attention to the task of unsupervised cross-domain person Re-ID.
in which the goal is to adapt the information learned from the labelled source domain to the unlabelled target domain.
In this case, both labelled datasets (source domain) and the easy-collected unlabelled dataset (target domain) can be used simultaneously.

Some domain transfer works~\cite{wei2018person,deng2018image,Zhong_2018_ECCV} have spent much
effort to solve the domain-shift problem.
One direct approach is to decrease  the domain gaps using 
the Generative Adversarial Network~\cite{goodfellow2014generative}.
Particularly, PTGAN~\cite{wei2018person} and SPGAN~\cite{deng2018image} aim to generate high-quality images in which image styles are transferred from source domain to target domain. 
At the same time, person identities are kept to ensure that the transferred images are satisfied for model training. 
Alternative methods~\cite{wang2018reid,huang2018eanet} are to discover the shared common knowledge between source and target domain to help the model training.
For instances, TJ-AIDL~\cite{wang2018reid} proposes a joint attribute-identity transfer learning to transfer the attribute information learned from labelled source data to unlabelled target data. 
EANet~\cite{huang2018eanet} fuses the human semantic information for both source and target data to provide additional supervision so as to narrow down the domain gap.
The drawback of these methods is that they all highly depend on the quality of the auxiliary knowledge, which is determined by other methods.

On the other hand, many works~\cite{li2018unsupervised,peng2016unsupervised,liu2017stepwise,song2018unsupervised,zhang2019self,fu2019self} proposes to generate reliable pseudo identity labels for the unlabelled target dataset.
In particular, authors of \cite{song2018unsupervised,zhang2019self,fu2019self} propose to use density-based clustering methods~\cite{campello2013density,ester1996density} for label estimation, which is proved to largely improve the pseudo-label quality.
Besides, the work of \cite{zhang2019self,fu2019self} also uses evenly divided part regions as additional representation to achieve high performance. 
Since these methods can not fully consider the subtle details and the whole data distribution in the target domain,
the performance is often still unsatisfactory in practice. 

\noindent\textbf{Memory Network.}
Memory-based learning network has been applied on various tasks~\cite{MemoryNetworks2015,graves2014neural,vinyals2016matching,sukhbaatar2015end,santoro2016meta,santoro2016one,Gong_2019_ICCV,xiong2016dynamic,zhong2019invariance,chandar2016hierarchical,wu2018unsupervised}. 
Among these existing methods, authors of \cite{chandar2016hierarchical} introduce a hierarchical memory structure to speed up memory access, in which memory cells are clustered to several groups, and then these groups are further organized to higher-level groups.
The difference from \cite{chandar2016hierarchical} is that we take advantage of multiple information from multi-level memory modules simultaneously
with respective supervision rather than one supervision for all levels.
Moreover, \cite{zhong2019invariance,wu2018unsupervised} construct an exemplar memory bank to store the instance-level feature for each image and then discover $k$-nearest neighbors based on it.
Compared with them, we explore the data relationships by considering different-level information comprehensively so that the feature representations are further improved with both fine-grained information and the global structure.
Besides, these memory modules complement each other to extract more reliable data relationships.

\section{The Proposed Method}
In the unsupervised cross-domain person Re-ID problem, we are given a labelled source training dataset $\{\left(x^i_s, y^i_s\right) \}$ with a set of person images $x_s^i$ and corresponding identity labels $y_s^i$ and an unlabelled target training dataset $\{x^i \}^{N}_{i=1}$ containing $N$ unlabelled images. The identity information in the target dataset is unavailable. 
There is usually a large domain gap between source and target data. The model trained on the labelled source dataset thus cannot be directly used on the target domain. 
Our goal is to learn discriminative embedding for the target domain by using both labelled source data and unlabelled target data. 

\subsection{Overview}
The proposed multi-level memory network (MMN) contains two main branches for training on labelled source domain and unlabelled target domain, respectively. The proposed multi-level memory module is applied to handle the unsupervised training on target domain.
As shown in Figure~\ref{fig:model}, the two branches share a same CNN backbone for feature extraction. Given an image, the backbone first extracts global-area feature map and then partitions it horizontally to two part feature maps, \ie, upper and bottom part as in \cite{fu2019self}. Following global average polling (GAP), $L_2$-normalization and ReLU activation, the model obtains $D$-dimensional embedding features for global area $\bbf_{\text{g}} \!\in\! \mbR^D$ and part areas $\mathbf{f}_{\text{pu}} \!\in\! \mbR^D$ and $\mathbf{f}_{\text{pb}} \!\in\! \mbR^D$, respectively.
In mini-batch training, half of each batch is from source domain and the other half is from target domain.

\par
Based on available identity labels,
we apply a simple classifier (based on fully-connected layer and softmax) on $\bbf_\text{g}$ following a cross-entropy loss to perform supervised learning in the source domain branch, which is denoted as $L_{\text{s}}$.
In the target domain branch,
$\bbf_\text{g}$, $\bbf_\text{pu}$ and $\bbf_\text{pb}$ are used
as the input of the multi-level module to obtain the similarity measurement.
As shown in Figure \ref{fig:model}, two loss functions perform directly based on the instance-level and domain-level memory, 
while the part-level memory helps to rectify the similarity measurement from instance level. 
Each memory is associated with a reading operation and a writing operation. 
Note that the reader of domain-level memory can provide a guidance for instance-level and part-level memory reading with a domain similarity measurment and soft assignment weights, while the part-level memory can rectify the memory read-out at instance level, as shown in Figure \ref{fig:model}. 
Details for each memory module are in the following sections. 

\subsection{Instance-level Memory Module}\label{sec:image}
We first define the instance-level memory as a matrix $\bMI\!\in\! \mbR^{D\times N}$ to
store the $D$-dimensional global image features of all $N$ sample. Let $\bm_{\tI,i}, \forall i \in [N]$ denote the $i$-th column of $\bMI$, where $[N]$ denotes the set of integers from $1$ to $N$. Each $\bm_{\tI,i}$ denotes a memory item corresponding to global-area feature $\bbf_{\text{g},i}$ of $i$-th target image $x_i$. 
All $\mathbf{m}_{\text{I},i}$ in $\mathbf{M}_{\text{I}}$ are initialized as zero vectors and dynamically updated via the writing operation. 

Given an $\bbf_{\text{g},i}$, the model can read out the relationships between the $i$-th sample and all others by measuring the similarity with all memory items $\bm_{\text{I}, j}$, which are used to obtain training supervisions in the target domain. To easily apply the similarity relationship in a classification scheme, we define the unique index of each sample as an individual class. 
The $i$-th image is assigned as class $i$. In this way, the similarity measurements can also be seen as the probability for classification.

\textbf{Reading Operation.}
To obtain the similarity relationships, we calculate the cosine similarities between $\bbf_{\text{g},i}$ and all $\bm_{\text{I}, j}$. The predicted probability is then obtained via a softmax function. The detailed formula is as follows:
\begin{equation}
\setlength{\abovedisplayskip}{0.2cm}
\setlength{\belowdisplayskip}{0.2cm}
    P_{\text{I}}(x_i,j)=\frac{\exp ( d ( \bbf_{\text{g},i}, \bm_{\text{I}, j} ) )}{{\sum}^{N}_{n=1} \exp ( d ( \bbf_{\text{g},i}, \bm_{\text{I}, n} ) )} ,
\label{eq:probability}
\end{equation}
where $d(\cdot,\cdot)$ is the cosine similarity measurement, formulated as
$d(\mathbf{u},\mathbf{v})=(\mathbf{u}\cdot \mathbf{v})/(||\mathbf{u}||\cdot ||\mathbf{v}||)/\alpha_1$.
$\alpha_1 \in (0,1]$ is a temperature fact scaling the distribution.

Based on the fact that there always exist same-identity samples in the target dataset~\cite{zhong2019invariance}, we assume that $x_i$ belongs to all the classes of the most relevant $k$ samples. 
We denote $\cK$ as the index set of the selected $k$ memory slots.
Then a multi-class objective for a mini-batch of size $B$ associated with the instance-level memory can be formulated as follows:
\begin{equation}
\setlength{\abovedisplayskip}{0.2cm}
\setlength{\belowdisplayskip}{0.2cm}
L_{\text{I}} = - \sum_{i=1}^{B} \sum_{j=1}^{N} \mu_{i,j} \log P_{\text{I}}(x_i, j),
\label{eq:imageloss}
\end{equation}
where $\mu_{i,j}$ denotes hard assignment weight. 
If $j\in \cK$, $\mu_{i,j}=1$; otherwise, $\mu_{i,j}=0$.
Here, $\cK$ can be generated from the top-$k$ similar samples in Eq.~\eqref{eq:probability} as in \cite{zhong2019invariance}.
However, we propose a new selection method guided by the domain-level memory in Section~\ref{sec:domain}.

\textbf{Write Operation.}
To update the memory slot, we write $\bbf_{\text{g},i}$ into the corresponding $i$-th memory slot by running average operation that is performed as:
\begin{equation}
\setlength{\abovedisplayskip}{0.2cm}
\setlength{\belowdisplayskip}{0.2cm}
    \mathbf{m}_{\text{I},i}\leftarrow \rho \mathbf{m}_{\text{I},i} + (1-\rho)\bbf_{\text{g},i},
\label{eq:imagewrite}
\end{equation}
where the coefficient $\rho$ represents the degree of the memory update.
A smaller $\rho$ indicates paying more consideration on the up-to-date feature, while the larger one on the representation saved in the memory slot.
After the updating, we re-normalize $\mathbf{m}_{\text{I},i}$ via $L_2$-normalization.

\subsection{Domain-level Memory Module}\label{sec:domain}
Although we have explored the relationships among individual images, 
it is still hard to fully capture the overall structure of the data since only $k$ samples are utilized.
We thus design a domain-level memory to store the representative prototypes of the whole target domain.

In detail, we first calculate the similarity metric $\mathbf{S}$ on all target data points via $k$-reciprocal encoding~\cite{zhong2017re}, which has been proved to be useful for Re-ID~\cite{zhang2019self,fu2019self,song2018unsupervised}. Then we utilize the clustering algorithm~\cite{campello2013density} to split the data points into $C$ pseudo-classes. 
The domain-level memory is designed to store the $D$-dimensional domain-level features of all pseudo-classes, defined as a matrix $\bMD\!\in\! \mbR^{D\times C}$.
The memory item $\mathbf{m}_{\text{D},c},\forall c\in[C]$ represents the representative feature of the $c$-th pseudo cluster, which is initialized as the feature of the $c$-th cluster centroid calculated by averaging all $\mathbf{f}_g$ of its elements.
$[C]$ denotes the set of integers from $1$ to $C$.
Note that $\bMD$ is a dynamic memory module since
$C$ changes every time calculating $\mathbf{S}$.

\textbf{Read Operation.}
Given an $\bbf_{\text{g},i}$ of target sample $x_i$ with pseudo class $c_i$, we can also obtain the predicted probability that $\bbf_{\text{g},i}$ belongs to each pseudo class via a softmax function, 
which is shown as:
\begin{equation}
\setlength{\abovedisplayskip}{0.2cm}
\setlength{\belowdisplayskip}{0.2cm}
    P_{\text{D}}(x_i, j)=\frac{\exp ( d ( \bbf_{\text{g},i}, \mathbf{m}_{\text{D},j} ) )}{{\sum}^{C}_{c=1} \exp ( d ( \bbf_{\text{g},i}, \mathbf{m}_{\text{D},c} ) )} ,
\label{eq:domain_probability}
\end{equation}
where $d(\cdot,\cdot)$ is the cosine similarity measurement which is the same as Eq.~\eqref{eq:probability}.

Based on Eq.~\eqref{eq:domain_probability}, we apply the cross-entropy loss as the objective function. Besides, we also use batch-hard triplet loss~\cite{hermans2017defense} to mine the relationships among image samples.
The final formula is shown as:
\begin{equation}
\setlength{\abovedisplayskip}{0.2cm}
\setlength{\belowdisplayskip}{0.2cm}
L_{\text{D}} = - \sum_{i=1}^{B} \log P_{\text{D}}(x_i, c_i) + L_{\text{Tri}},
\label{eq:domainloss}
\end{equation}
where $L_{\text{Tri}}$ denotes the batch-hard triplet loss~\cite{hermans2017defense}.

\textbf{Write Operation.}
In the write process of the domain-level memory, we only update the $c_i$-th memory item, \ie, the class centroid of itself. The update process is formulated as follows:
\begin{equation}
\setlength{\abovedisplayskip}{0.2cm}
\setlength{\belowdisplayskip}{0.2cm}
    \mathbf{m}_{\text{I},c_i}\leftarrow \rho \mathbf{m}_{\text{I},c_i} + (1-\rho)\bbf_{\text{g},i},
\label{eq:domainwrite}
\end{equation}
where $\rho$ is the same coefficient as in Eq.~\eqref{eq:imagewrite}. We also use $L_2$-normalization on $\mathbf{m}_{\text{I},c_i}$ after updating.

\textbf{Domain-level guidance for memory read.}
Aparat from the training objective above, the domain-level memory also provides guidance for the memory read on the instance level. 
In Eq.~\eqref{eq:imageloss}, we select $\cK$ with $k$ samples relying on the cosine similarity in Eq.~\eqref{eq:probability}, which have limited confidence especially in the early training stage due to the unreliable features. 
To address the problem, we propose a new selection method for $\cK$ in the instance level guided by the high-level information in the domain level.
Specially, given a target sample $x_i$, we first extract the most similar $2k$ samples based on $\mathbf{S}$ and do the same thing for each selected sample.
We assume that if a sample A is more similar to B than C, the number of the overlapped samples in the selected $2k$ samples of A and B is larger than A and C. If B and C have the same number of overlapped samples with A, $S_{A,B}$ and $S_{A,C}$ (\ie the values in the similarity matrix $\mathbf{\text{S}}$) will be used to measure the similarity. 
According to this, the selected $2k$ samples can be reordered.
After that, we select the top-$k$ samples based on the reordered rank and record their indexes in a set as  $\widetilde{\cK}$, will be used for the operations on the instance level and part level, reflecting the guidance from the domain-level memory. 

\par
Moreover, if A is more similar to B than C, it is more likely that A and B belong to same identity, and it is reasonable to assign larger weight on B when applying the training loss Eq.~\eqref{eq:imageloss} on A.
We thus define a similarity-based soft assignment weight on $\widetilde{\cK}$ relying on the similarity matrix $\mathbf{S}$:
\begin{equation}
\setlength{\abovedisplayskip}{0.2cm}
\setlength{\belowdisplayskip}{0.2cm}
    w_{i,j}=\text{exp}(-\alpha_2 (1-S_{i,j})), \forall j \in \widetilde{\cK}, 
\label{eq:softweight}
\end{equation}
where $\alpha_2$, $\alpha_2>0$ is a temperature parameter to control the importance of the selected samples.
Since $S_{i,j}\in (0, 1]$, the soft weight $w_{i,j}\in (0, 1]$.
By applying Eq.~\eqref{eq:softweight}, Eq.~\eqref{eq:imageloss} can be modified as:
\begin{equation}
\setlength{\abovedisplayskip}{0.1cm}
\setlength{\belowdisplayskip}{0.1cm}
L_{\text{I}} = - \sum_{i=1}^{B} \sum_{j=1}^{N} (\mu_{i,j}\cdot w_{i,j}) \log P_{\text{I}}(x_i, j),
\label{eq:imageloss2}
\end{equation}
where if $j\in \widetilde{\cK}$, $w_{i,j}$ is defined by Eq.~\eqref{eq:softweight}; if $j\notin \widetilde{\cK}$, $w_{i,j}=0$. 

\par
With the guidance from the domain-level memory, we can select more confident top-$k$ samples $\cK$ for the instance-level training objective as well as assign different weights for them according to the higher level similarity measurement. 

\subsection{Part-level Memory Module}\label{sec:part}
As illustrated in Figure~\ref{fig:figure1}, it is important to consider the subtle differences of different identities since their overall appearances may look similar to each other. 
In particular, we design a part-level memory $\mathbf{M}_\text{P}$ consisting of two components for memorizing the information from upper body and bottom body, denoted as $\mathbf{M}_\text{PU}$ and $\mathbf{M}_{\text{PB}}$ respectively. 

Given an upper-body feature $\bbf_{\text{pu},i}$ of target sample $x_i$, we also obtain the predicted probability using the same process in Eq.~\eqref{eq:probability}:
\begin{equation}
\setlength{\abovedisplayskip}{0.2cm}
\setlength{\belowdisplayskip}{0.2cm}
    P_{\text{PU}}(x_i,j)=\frac{\exp ( d ( \bbf_{\text{pu},i}, \bm_{\text{PU}, j} ) )}{{\sum}^{N}_{n=1} \exp ( d ( \bbf_{\text{pu},i}, \bm_{\text{PU}, n} ) )} ,
\label{eq:part_probability}
\end{equation}
where $\bm_{\text{PU}, j}$, $j\in [N]$ denotes the $j$-th column of $\mathbf{M}_\text{PU}$.
$d(\cdot,\cdot)$ is the same as Eq.~\eqref{eq:probability}.
The predicted probability $P_{\text{PB}}(x_i,j)$ for bottom body is obtained in the same way.

\textbf{Part-level rectification for memory read-out.}
We utilize the part-level information from Eq.~\eqref{eq:part_probability} as the complementary guidance to rectify the soft weight in Eq.~\eqref{eq:softweight} for the instance-level memory.
To be specific, we first apply $0$-$1$ normalization on $P_{\text{PU}}(x_i,j)$ and $P_{\text{PB}}(x_i,j)$ to force the value of each element in the range of $0$ to $1$.
The soft weight in Eq.~\eqref{eq:softweight} is then modified as:
\begin{equation}
\setlength{\abovedisplayskip}{0.2cm}
\setlength{\belowdisplayskip}{0.2cm}
    w_{i,j}  \leftarrow  (1-\gamma) w_{i,j} + \gamma (P_{\text{PU}}(x_i,j) + P_{\text{PB}}(x_i,j)),
\label{eq:softweight2}
\end{equation}
where $\gamma$ reflects the degree of the rectification by the part information. We apply Eq.~\eqref{eq:softweight2} into Eq.~\eqref{eq:imageloss2} to build the final objective $L_\text{I}$ in the instance-level memory.

\subsection{Training Loss}
By combining Eq.~\eqref{eq:softweight2}, Eq.~\eqref{eq:imageloss2}, Eq.~\eqref{eq:domainloss} in the target domain and $L_{\text{s}}$ in the source domain together, we construct a joint optimization function for our MMN formulated as:
\begin{equation}
\setlength{\abovedisplayskip}{0.2cm}
\setlength{\belowdisplayskip}{0.2cm}
    L = (1-\lambda) L_{\text{s}} + \lambda(L_{\text{I}} + \beta L_{\text{D}}),
\label{eq:totalloss}
\end{equation}
where $\lambda$ and $\beta$ are loss weights. $\lambda$ controls the proportion of the source loss and and the target loss. $\beta$ is responsible for measuring the importance of domain-level loss.

\section{Experiments}
\subsection{Datasets and Evaluation Protocol}

\textbf{Datasets.} We evaluate the proposed MMN on three large-scale person Re-ID datasets.

\textit{Market-1501}~\cite{bai2017scalable} includes 32,668 labelled images of 1,501 identities from 6 cameras. All the pedestrians are detected using DPM detector~\cite{felzenszwalb2009object}. There are 12,936 images of 751 identities for the training set and 19,732 images of 750 identities for the test set.

\textit{DukeMTMC-Re-ID}~\cite{zheng2017unlabeled} comprises 36,411 labelled images belonging to 1,404 identities collected from 8 camera viewpoints. 
It is divided into the training set with 16,522 images of 702 identities and the test set with the remaining 19,889 images of 702 identities. 
For simplicity, we use the term ``Duke'' to represent this dataset.

\textit{MSMT17}~\cite{wei2018person} is a newly released dataset consisting of 126,441 images from 4,101 identities. The dataset is captured by 15 camera views, in which 12 cameras are outdoor, and 3 are indoor. 
The person detector is Faster RCNN~\cite{ren2015faster}.
32,621 images of 1,041 identities are used as the training set, while 93,820 images of 3,060 identities as the test set. 

\textbf{Evaluation Protocol.} 
We use the cumulative match characteristic (CMC) curve~\cite{gray2007evaluating} and the mean average precision (mAP)~\cite{bai2017scalable} as the evaluation metrics.
The CMC reflects the match scores that a query is in the various sizes of candidate lists. We report the Rank-1 score to represent the CMC curve. 
For each query image, the average precision (AP) is first computed from its precision-recall curve, and the mAP is then calculated as the mean value of APs of all queries. 
Note that we fairly report all the results under the single-shot setting as~\cite{zhang2019self,fu2019self}, and there are no other post-processing methods like re-ranking~\cite{zhong2017re}.

\begin{table}[t!]
\small
\setlength{\belowcaptionskip}{-0.1cm}
\setlength{\abovecaptionskip}{-0.1cm}
\begin{center}
\setlength{\tabcolsep}{1.48mm}{
\begin{tabular} {l|ccc|c|c|c|c}
\hline
\multirow{2}{*}{Method} & \multicolumn{3}{c|}{Module} & \multicolumn{2}{c|}{D$\rightarrow$M}      & \multicolumn{2}{c}{M$\rightarrow$D} \\ 
\cline{2-8} 
& $\mathbf{M}_\text{I}$ & $\mathbf{M}_\text{D}$ & $\mathbf{M}_\text{P}$ & mAP & Rank-1 & mAP & Rank-1 \\ 
\hline
\hline
Baseline  & - & - & - & 17.7 & 43.7 & 12.9 & 27.4 \\ 
\hline
MMN & $\checkmark$ & - & - & 50.3 & 79.5 & 48.3 & 68.9 \\
MMN & $\checkmark$ & - & $\checkmark$ & 55.2 & 81.3 & 49.8 & 70.0 \\
MMN & $\checkmark$ & $\checkmark$ & - & 60.3 & 84.4 & 52.4 & 72.4 \\
MMN & $\checkmark$ & $\checkmark$ & $\checkmark$ & \textbf{65.1} & \textbf{86.0} & \textbf{53.9} & \textbf{73.2} \\
\hline
\end{tabular}}
\end{center}
\caption{Effectiveness of the multi-level memory module. $\mathbf{M}_\text{I}$, $\mathbf{M}_\text{D}$ and $\mathbf{M}_\text{P}$ represent the instance-level, domain-level and the part-level memory module,  respectively. D$\rightarrow$M means that we use Duke~\cite{zheng2017unlabeled} as the source domain and Market-1501~\cite{bai2017scalable} as the target domain. Baseline means directly using the model trained on source domain to the target domain. 
} 
\label{tab:Effectiveness of different part}
\end{table}

\begin{figure}[t!]
\centering
\includegraphics[trim =0mm 0mm 0mm 0mm, clip, width=0.493\linewidth]{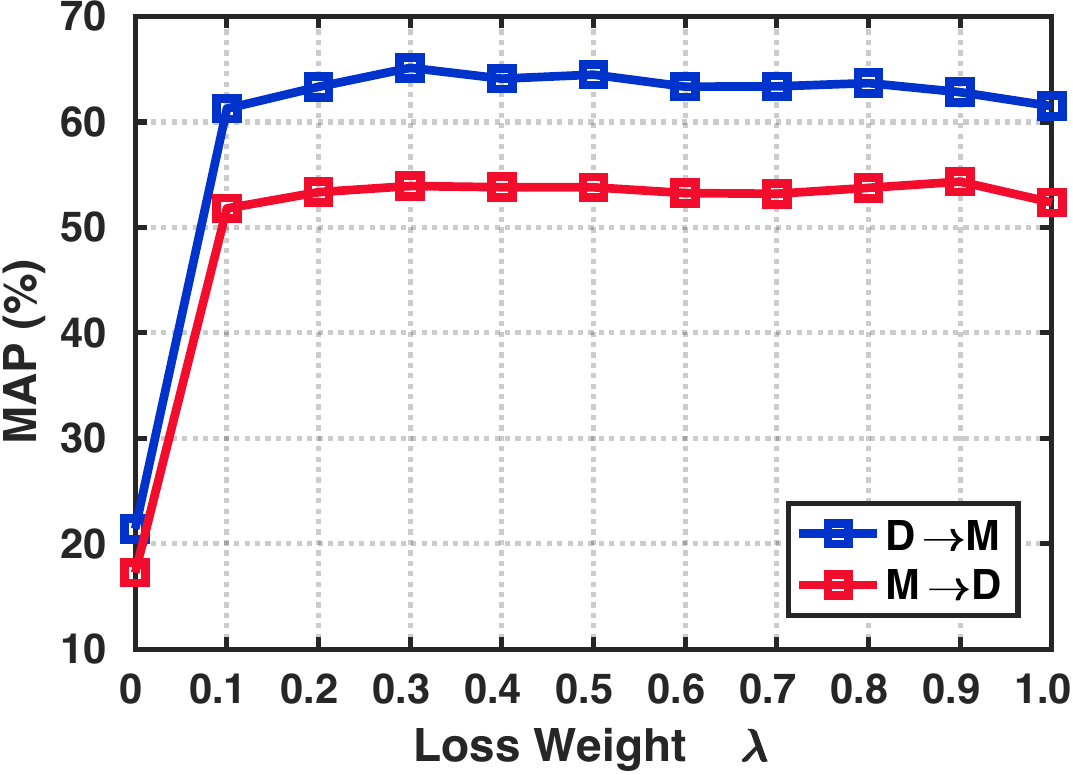} 
\includegraphics[trim =0mm 0mm 0mm 0mm, clip, width=0.493\linewidth]{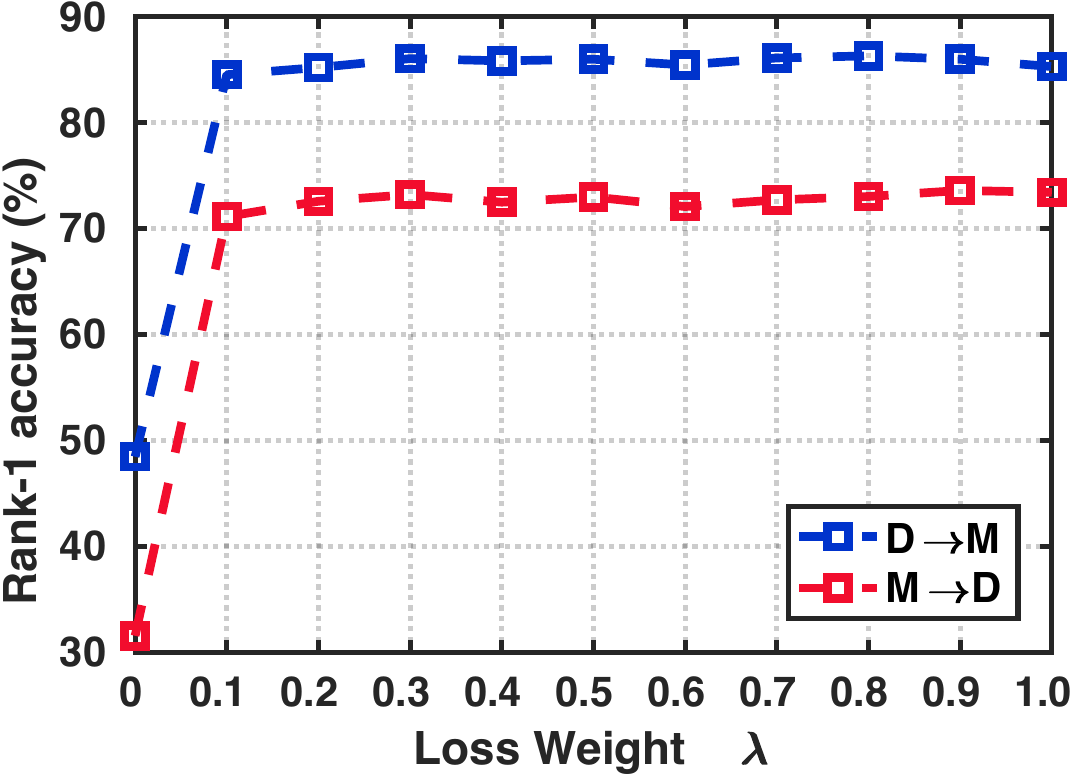} 
\setlength{\abovecaptionskip}{-0.2cm} 
\setlength{\belowcaptionskip}{-0.3cm}
\caption{Parameter analysis of the loss weight $\lambda$ on D$\rightarrow$M and M$\rightarrow$D.} 
\label{fig:lamda}
\end{figure}

\subsection{Implementation Details}
\textbf{Model.} 
We adopt ResNet-50~\cite{resnet50} as the backbone of our model without the last classification layer, which is pre-trained on ImageNet~\cite{deng2009imagenet}. 
Same as~\cite{zhang2019self,sun2018beyond,huang2018eanet}, the stride of the last residual block is set from $2$ to $1$.
Moreover, we fix the weights of the first two residual blocks to save GPU memory similar as~\cite{zhong2019invariance}. 

\textbf{Preprocessing.} 
All input images are resized to $256\times128$. 
Random flipping and random erasing~\cite{zhong2017random} are employed as the data augmentation for training stage. 
Same as~\cite{zhong2019invariance}, we use the generated camera-style images~\cite{zhong2019camstyle} for the unseen target domain to increase the image diversity.

\textbf{Training Settings.} 
Following~\cite{zhong2019invariance}, we set the mini-batch size to $128$ for both source and target images. 
All experiments use the SGD optimizer with a momentum of $0.9$ and a weight decay of $5\times10^{-4}$.
We train the model with an initial learning rate of $0.01$ for the backbone and $10$ times for the other layers. 
After $40$ epochs, the learning rate is divided by $10$. 
The total epochs of the training stage are $60$.
The coefficient $\rho$ is set to $\rho=0.01\times \text{epoch}$.
Note that the domain-level memory is added at the $10$-th epoch.
We set the number of the selected samples $k=10$ and change $\mathbf{S}$ every $2$ epochs. 
Without otherwise notation, we set $\alpha_1=0.05$, $\alpha_2=2.0$.

For evaluation, we concatenate $\bbf_{\text{g}}$, $\bbf_{\text{pu}}$ and $\bbf_{\text{pb}}$ as the feature representation.
When removing the part-level memory, we only use the global-area feature embedding.
Cosine similarity is used as the evaluation metric.
Specially, we denote the model trained with an identity classifier without any memory components as the \textit{baseline}.

\begin{figure}[t!]
\centering
\includegraphics[trim =0mm 0mm 0mm 0mm, clip, width=0.493\linewidth]{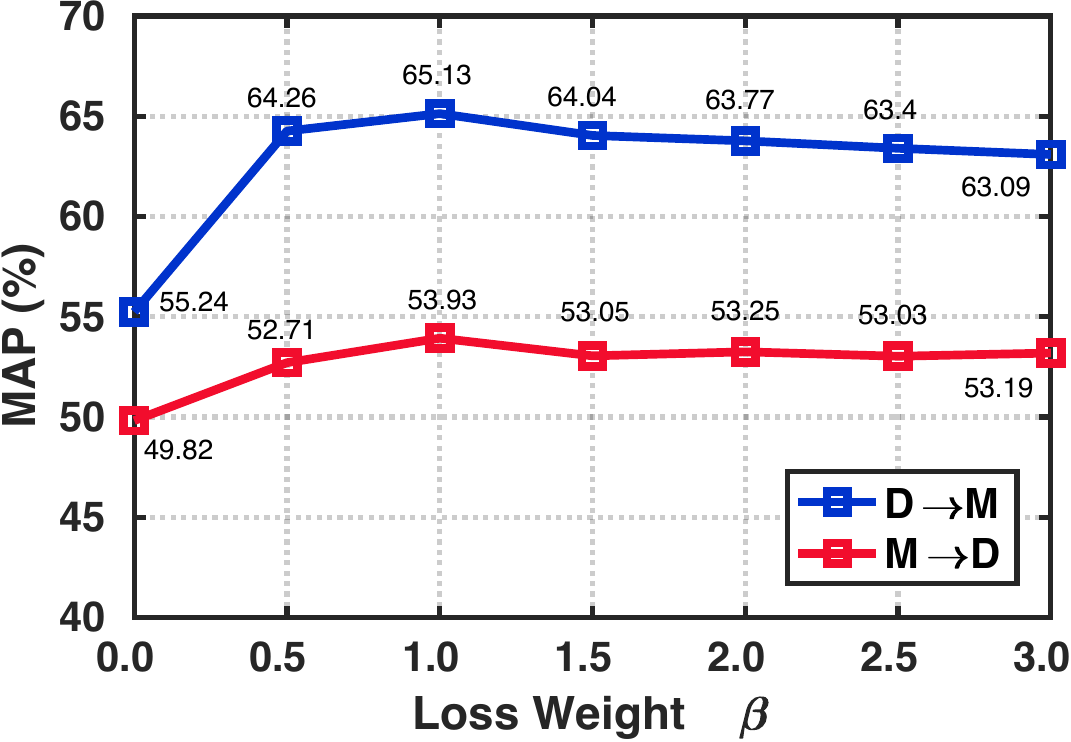} 
\includegraphics[trim =0mm 0mm 0mm 0mm, clip, width=0.493\linewidth]{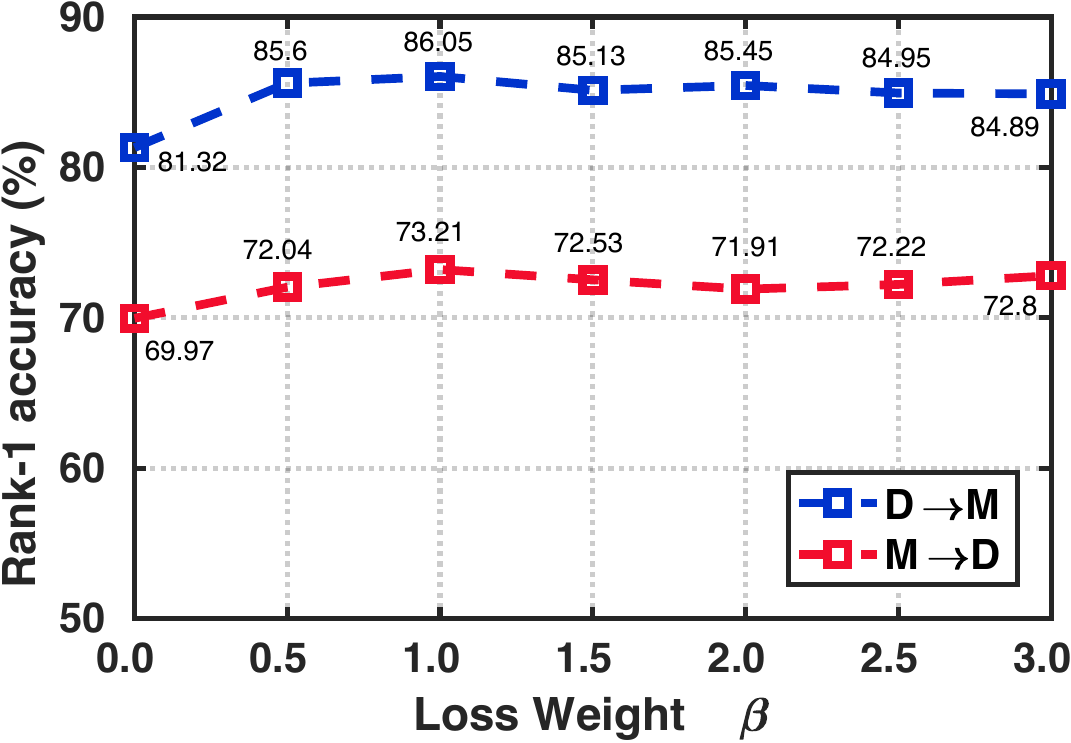} 
\setlength{\abovecaptionskip}{-0.2cm} 
\setlength{\belowcaptionskip}{-0.2cm}
\caption{Parameter analysis of the loss weight $\beta$ on D$\rightarrow$M and M$\rightarrow$D.} 
\label{fig:beta}
\end{figure}

\begin{figure}[t!]
\centering
\includegraphics[trim =0mm 0mm 0mm 0mm, clip, width=.975\linewidth]{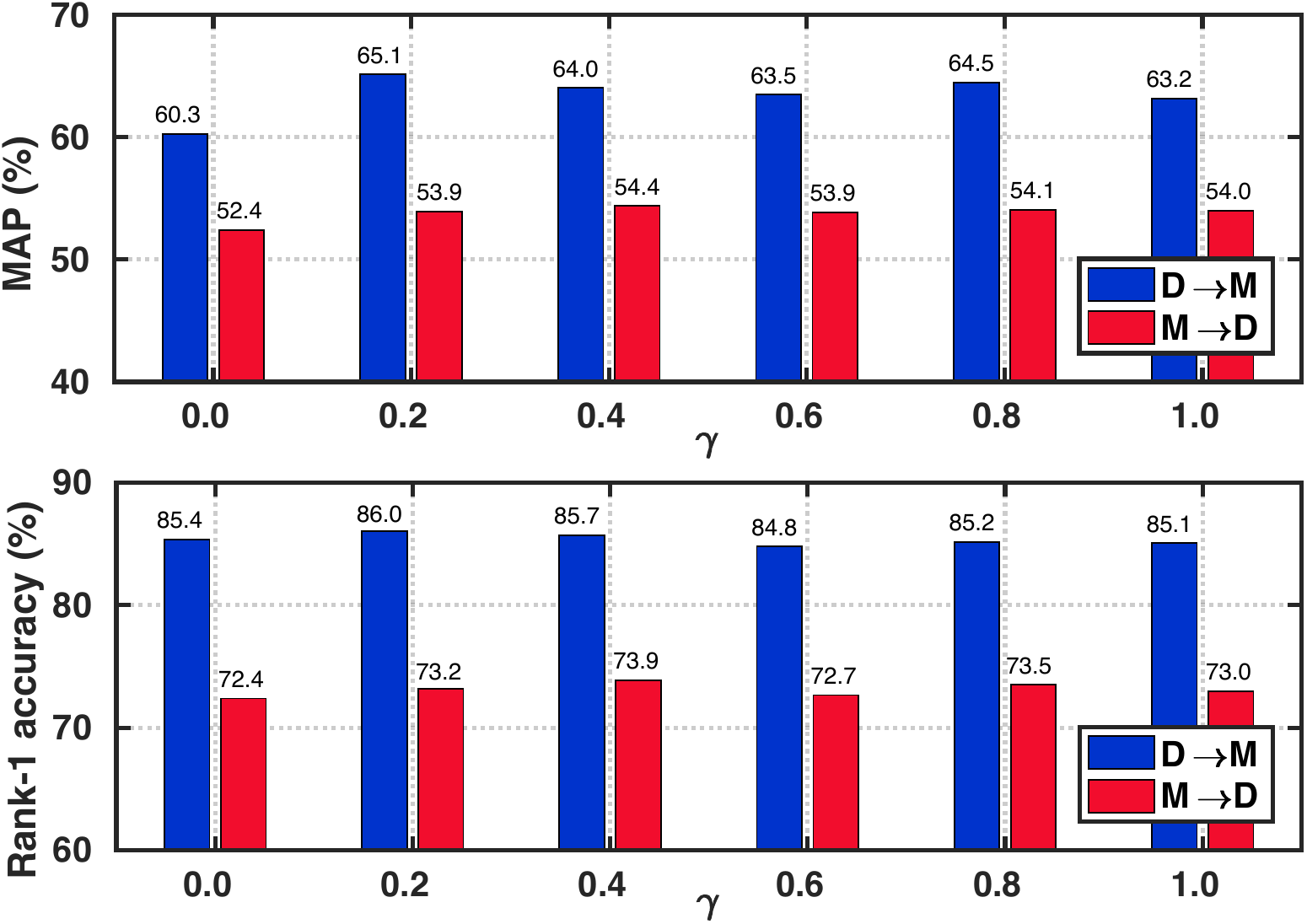} 
\setlength{\abovecaptionskip}{0.1cm} 
\setlength{\belowcaptionskip}{-0.6cm}
\caption{Parameter analysis of the degree of the soft-weight rectification $\gamma$ on D$\rightarrow$M and M$\rightarrow$D.} 
\label{fig:gamma}
\end{figure}

\begin{figure*}[t!]
\centering
\includegraphics[trim =0mm 0mm 0mm 0mm, clip, width=.95\linewidth]{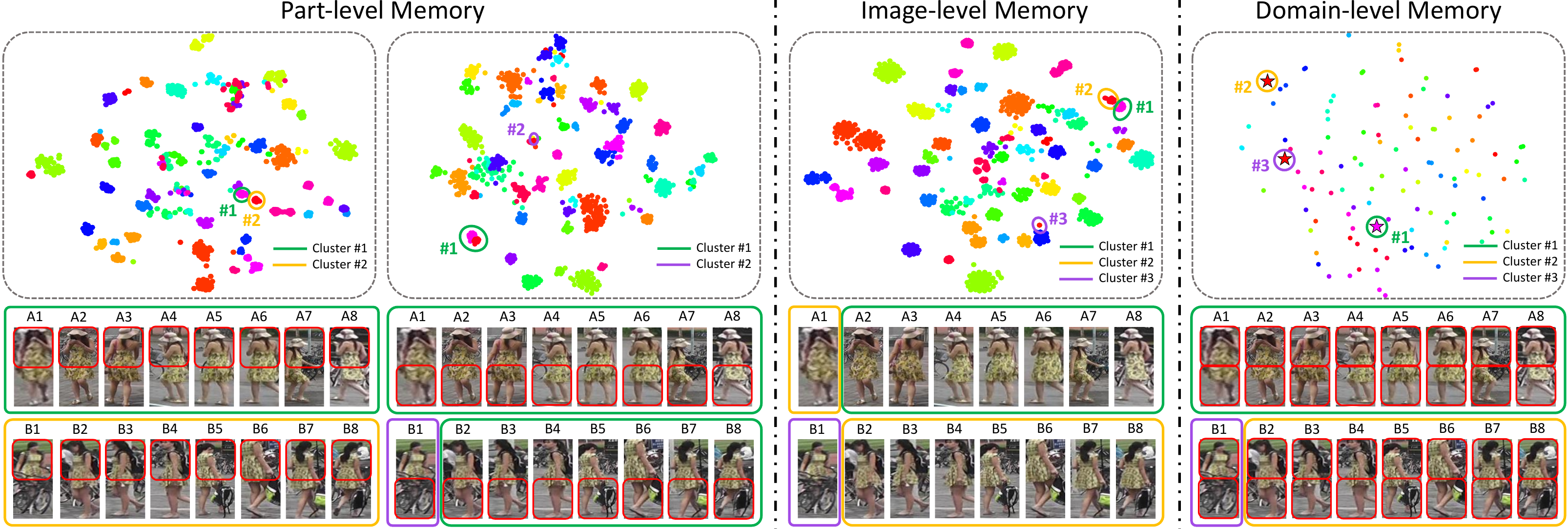} 
\setlength{\abovecaptionskip}{0.13cm} 
\setlength{\belowcaptionskip}{-0.3cm}
\caption{Illustration of the effectiveness of multi-level memory module. Images in the fist row are drawn from t-SNE~\cite{van2014accelerating} which reflects the distribution of the memory slots in different level. The second-row pictures represent the example images of two identities A and B. For part-level and instance-level memory, we use clustering method \cite{campello2013density} on the memory slots to generate the pseudo labels.
The \textcolor[rgb]{0.2,0.7,0.3}{green},
\textcolor[rgb]{1.0,0.8,0.1}{yellow},
and \textcolor[rgb]{0.6,0.0,1.0}{purple} rectangles shows different clusters.
Memory in different level provides multiple representations for the images from fine-grained information to global data structure.
}
\label{fig:visualization}
\end{figure*}

\subsection{Ablation Study}
In this section, we first perform several ablation studies to investigate the importance of each memory module in our MMN.
Then we conduct series experiments to analyze the sensitivities of some important hyper-parameters.

\begin{table}[t!]
\small
\setlength{\belowcaptionskip}{-0.1cm}
\setlength{\abovecaptionskip}{-0.4cm}
\begin{center}
\setlength{\tabcolsep}{1.6mm}{
\begin{tabular}{l|c|c|cc}
\hline
\multirow{2}{*}{Method} & \multicolumn{2}{c|}{D$\rightarrow$M} & \multicolumn{2}{c}{M$\rightarrow$D} \\ 
\cline{2-5} 
 & mAP & Rank-1 & \multicolumn{1}{c|}{mAP} & Rank-1 \\ 
\hline
\hline
MMN (w/ $\mathbf{M}_\text{I}$) w/o G & 45.8 & 78.0 & \multicolumn{1}{c|}{40.3} & 63.3 \\
MMN (w/ $\mathbf{M}_\text{I}$) & 50.3 & 79.5 & \multicolumn{1}{c|}{48.3} & 68.9   \\ 
\hline
MMN w/o G & 60.8 & 82.5  & \multicolumn{1}{c|}{51.2} & 70.9   \\
MMN & \textbf{65.1}  & \textbf{86.0} & \multicolumn{1}{c|}{\textbf{53.9}} & \textbf{73.2}   \\ 
\hline
\end{tabular}}
\end{center}
\caption{Effectiveness of domain-level guidance for memory read on D$\rightarrow$M and M$\rightarrow$D. G denotes the domain-level guidance. The w/ $\mathbf{M}_\text{I}$ means MMN reduces to only using $\mathbf{M}_\text{I}$ as in Table~\ref{tab:Effectiveness of different part}.}
\label{tab:rss}
\end{table}

\textbf{Effectiveness of Multi-level Memory Module.} 
We evaluate the effectiveness of each memory module by adding them gradually.
As shown in Table~\ref{tab:Effectiveness of different part}, only with the instance-level memory $\mathbf{M}_\text{I}$, we improve the performance by $32.6\%$ and $35.4\%$ in mAP compared with the baseline for D$\rightarrow$M and M$\rightarrow$D .
After adding the part-level memory $\mathbf{M}_\text{P}$ into the model, 
the mAP and Rank-$1$ accuracy further increase by $4.9\%$ and $1.8\%$ for D$\rightarrow$M and $1.5\%$ and $1.1\%$ for M$\rightarrow$D.
It shows that part-level guidance is important to improve the feature representation with fine-grained information.
Compared with only using $\mathbf{M}_\text{I}$, adding the domain-level memory $\mathbf{M}_\text{D}$ can provide $10.0\%$ and $4.1\%$ mAP improvement for D$\rightarrow$M and M$\rightarrow$D respectively.
This demonstrates that $\mathbf{M}_\text{D}$ has a great impact on the model generalization by considering the global structure in the target domain.
Moreover, when integrating $\mathbf{M}_\text{I}$, $\mathbf{M}_\text{D}$ and $\mathbf{M}_\text{P}$ together, our multi-level module network (MMN) gain a significant improvement in performance.
For instance, MMN achieves $65.1\%$ and $53.9\%$ in mAP for D$\rightarrow$M and M$\rightarrow$D, which are $14.8\%$ and $5.6\%$ higher than only using $\mathbf{M}_\text{I}$.

In addition, we provide a visualization in Figure~\ref{fig:visualization} to intuitively see whether the memory modules have learned effective information after training.
It is shown that $\mathbf{M}_\text{I}$ provides more fine-grained details for images while $\mathbf{M}_\text{D}$ focuses on the global distribution in the whole dataset.

\begin{table}[t!]
\small
\setlength{\abovecaptionskip}{-0.4cm}
\begin{center}
\setlength{\tabcolsep}{2.0mm}{
\begin{tabular}{l|c|c|cc}
\hline
\multirow{2}{*}{Method} & \multicolumn{2}{c|}{D$\rightarrow$M} & \multicolumn{2}{c}{M$\rightarrow$D} \\ 
\cline{2-5} 
 & mAP & Rank-1 & \multicolumn{1}{c|}{mAP} & Rank-1 \\ 
\hline
\hline
MMN w/o $\mathbf{M}_\text{D}$ & 55.2 & 81.3  & \multicolumn{1}{c|}{49.8} & 70.0   \\
MMN w/ FC & 62.6 & 84.6  & \multicolumn{1}{c|}{52.9} & 71.9   \\
MMN & \textbf{65.1}  & \textbf{86.0} & \multicolumn{1}{c|}{\textbf{53.9}} & \textbf{73.2}   \\ 
\hline
\end{tabular}}
\end{center}
\caption{Comparison of the memory module and an FC layer on the domain level. FC denotes the fully-connected layer. The w/o $\mathbf{M}_\text{D}$ means MMN reduces to only using $\mathbf{M}_\text{I}$ and $\mathbf{M}_\text{P}$ as in Table~\ref{tab:Effectiveness of different part}.}
\label{tab:memory}
\end{table}

\begin{table*}[t!]\footnotesize
\setlength{\abovecaptionskip}{-0.2cm}
\begin{center}
\begin{tabu} to 0.8971\textwidth {l|X[c]|X[c]|X[c]|X[c]|X[c]|X[c]|X[c]|X[c]}
\hline
\multicolumn{1}{l|}{\multirow{2}{*}{Methods}}&\multicolumn{2}{c|}{D$\rightarrow$M}&\multicolumn{2}{c|}{M$\rightarrow$D}&\multicolumn{2}{c|}{D$\rightarrow$MS}&\multicolumn{2}{c}{M$\rightarrow$MS}\\
\cline{2-9}
\multicolumn{1}{c|}{} & mAP & Rank-1 & mAP & Rank-1 & mAP & Rank-1 & mAP & Rank-1 \\
\hline
\hline
PTGAN \cite{wei2018person}'18 & - & 38.6 & - & 27.4 & 3.3 & 11.8 & 2.9 & 10.2 \\
PUL \cite{fan2017pul}'18 & 20.5 & 45.5 & 16.4 & 30.0 & - & - & - & - \\
SPGAN \cite{deng2018image}'18 & 22.8 & 51.5 & 22.3 & 41.1 & - & - & - & - \\
MMFA \cite{lin2018multibmvc}'18 & 27.4 & 56.7 & 24.7 & 45.3 & - & - & - & - \\
{SPGAN+LMP} \cite{deng2018image}'18 & 26.7 & 57.7 & 26.2 & 46.4 & - & - & - & - \\
TJ-AIDL \cite{wang2018reid}'18 & 26.5 & 58.2 & 23.0 & 44.3 & - & - & - & - \\
HHL \cite{Zhong_2018_ECCV}'18 & 31.4 & 62.2 & 27.2 & 46.9 & - & - & - & - \\
EANet~\cite{huang2018eanet}'19 & 51.6 & 78.0 & 48.0 & 67.7 & - & - & - & - \\
CamStyle \cite{zhong2019camstyle}'19 & 27.4 & 58.8 & 25.1 & 48.4 & - & - & - & - \\
DECAMEL \cite{yu2018unsupervised}'19 $^\dagger$& 32.4 & 60.2 & - & - & - & - & 11.1 & 30.3\\
MAR \cite{yu2019unsupervised}'19 $^\ddagger$ & 40.0 & 67.7 & 48.0 & 67.1 & - & - & - & - \\
SCAN \cite{chenself}'19& 30.4 & 61.0 & 28.4 & 48.4 & - & - & - & - \\
PAUL \cite{yang2019patch}'19& 36.8 & 66.7 & 35.7 & 56.1 & - & - & - & - \\
UDA \cite{song2018unsupervised}'19 & 53.7 & 75.8 & 49.0 & 68.4 & - & - & - & - \\ 
ECN \cite{zhong2019invariance}'19 & 43.0 & 75.1 & 40.0 & 63.3 & 10.2 & 30.2 & 8.5 & 25.3\\
PAST \cite{zhang2019self}'19& 54.5 & 78.4 & \textcolor{red}{\textbf{54.3}} & 72.4 & - & - & - & - \\
SSG \cite{fu2019self}'19& \textcolor{blue}{58.3} & \textcolor{blue}{80.0} & 53.4 & \textcolor{blue}{73.0} & \textcolor{blue}{13.3} & \textcolor{blue}{32.2} & \textcolor{blue}{13.2} & \textcolor{blue}{31.6} \\
\hline
Baseline & 17.7 & 43.7 & 12.9 & 27.4 & 4.5 & 15.0 & 2.6 & 8.9\\
\hline
MMN & \textcolor{red}{\textbf{65.1}} & \textcolor{red}{\textbf{86.0}} & \textcolor{blue}{53.9} & \textcolor{red}{\textbf{73.2}} & \textcolor{red}{\textbf{17.2}} & \textcolor{red}{\textbf{43.2}} & \textcolor{red}{\textbf{14.1}} & \textcolor{red}{\textbf{36.8}} \\
\hline
\end{tabu}
\end{center}
\caption{\label{tabel:sota} Performance comparison with state-of-the-art methods under the unsupervised cross-domain setting. \textbf{M}: Market-1501 \cite{zheng2015scalable}. \textbf{D}: DukeMTMC-reID \cite{zheng2017unlabeled}. \textbf{MS}: MSMT17 \cite{wei2018person}. A$\rightarrow$B represents that A (source dataset) transfers to B (target dataset). $\dagger$ denotes that the source domain is a combination of seven datasets while $\ddagger$ denotes MSMT17 is used as the source domain. We mark the \textcolor{red}{\textbf{1st}} and \textcolor{blue}{2rd} highest scores to \textcolor{red}{\textbf{red}} red and \textcolor{blue}{blue} respectively.}
\end{table*}

\textbf{Parameter Analysis for $\lambda$, $\beta$ and $\gamma$.} 
In Figure~\ref{fig:lamda}, we first analyze the loss weight $\lambda$ in Eq.~\eqref{eq:totalloss}.
When $\lambda=0$, only source dataset is used for training while the model reduces to the baseline model.
When applying MMN on the target domain, the performance can be improved consistently by a large margin whatever $\lambda$ is.
It means that our MMN is beneficial for the model generalization on the target domain.
Note that our MMN can produce a satisfactory result even $\lambda=1$, \ie, without source data.
We believe that it is benefited from our MMN learning discriminative features using multi-level information on the unlabelled target data.

We then compare different values of the loss weight $\beta$ in Eq.~\eqref{eq:totalloss} as shown in Figure~\ref{fig:beta}, which measures the degree of importance of the domain-level memory $\mathbf{M}_\text{D}$.
When $\beta=0$, the model reduces to MMN w/o $\mathbf{M}_\text{D}$ as in Table~\ref{tab:Effectiveness of different part}.
When $\beta>0$, the performance gains consistent improvement, which verifies the effectiveness of $\mathbf{M}_\text{D}$.
With increasing $\beta$ continually, the result would get a platform.

Furthermore, we explore the effect of the hyper-parameter $\gamma$ in Eq.~\eqref{eq:softweight2}, which is the degree of the rectification guided by $\mathbf{M}_\text{P}$.
When $\gamma=0$, the model is changed to MMN w/o $\mathbf{M}_\text{P}$ as in Table~\ref{tab:Effectiveness of different part}, while the soft assignment weight reduces from Eq.~\eqref{eq:softweight2} to Eq.~\eqref{eq:softweight}.
In Figure~\ref{fig:gamma}, we can see that our MMN achieves the improvement in performance after adding $\mathbf{M}_\text{P}$ into the training when $\gamma>0$.
It is worth noting that our approach can still improve the mAP even only using the probability from the part-level memory as the soft weight, \ie, when $\gamma=1.0$ in Eq.~\eqref{eq:softweight2}.
This demonstrates the importance of the fine-grained information from the part-level memory.

To sum up, we set $\lambda=0.3$, $\beta=1.0$ and $\gamma=0.2$ in the next experiments.

\textbf{Effectiveness of Domain-level Guidance for Memory Read.} 
In Table~\ref{tab:rss}, we evaluate the importance of domain-level guidance for memory read, \ie, top-$k$ similar sample selection with soft assignment weight in Eq.~\eqref{eq:softweight}.
If the model is trained without this guidance, the sample selection and the weights for different samples are reduced to the traditional method, as described in Eq.~\eqref{eq:imageloss}.
We observe that our MMN largely improves the performance with the guidance of the domain-level memory by $4.3\%$ and $2.7\%$ in mAP and $3.5\%$ and $2.3\%$ in Rank-$1$ for D$\rightarrow$M and M$\rightarrow$D respectively.
When the model is only trained with the instance-level memory, the performance can still improve, especially for M$\rightarrow$D.
This shows that our method can select more reliable $k$ nearest samples with domain-level guidance.
Meanwhile, it also shows that it is important to consider the soft weight for each sample in the $k$ neighbors, in which the weight guidance is from the similarity relationships generated from the domain level.

\textbf{Comparison of Domain-level Memory with FC Layer.} 
In order to verify the effectiveness of the domain-level memory, we build a fully-connected layer (FC) with the softmax cross-entropy loss as the classifier in the domain level.
For fairness, we also initialize the FC by calculating the mean embedding features for each cluster.
Table~\ref{tab:memory} shows that both the FC-based classifier and the memory-based classifier can get large improvement compared with the model without domain level. It proves that global-structure information is useful for model generalization.
Meanwhile, our MMN achieves better results than the FC-based classifier, which validates the advantage of the memory module.

\subsection{Comparison with the State-of-the-art Methods}
In Table~\ref{tabel:sota}, we compare our MMN with the state-of-the-art unsupervised methods following the evaluation setting in \cite{zhang2019self,zhong2019invariance,fu2019self}.
Our proposed MMN is competitive or superior to the previous methods. 
In particular, our MMN achieves $65.1\%/86.0\%$ for D$\rightarrow$M, $53.9\%/73.2\%$ for M$\rightarrow$D, $17.2\%/43.2\%$ for D$\rightarrow$MS and $14.1\%/36.8\%$ for M$\rightarrow$MS in mAP$/$Rank-$1$, which is higher than the relatively best existing method SSG~\cite{fu2019self} by $6.8\%/6.0\%$, $0.5\%/0.2\%$, $4.1\%/11.0\%$ and $0.9\%/5.2\%$ respectively.
The reason that our MMN has a little bit lower result than PAST~\cite{zhang2019self} on mAP for M$\rightarrow$D is that
PAST~\cite{zhang2019self} concatenates $9$ part features together and uses the pre-trained model from the source domain as the initialization.

\section{Conclusion}
In this paper, we present a multi-level memory network (MMN) for the unsupervised cross-domain person Re-ID task.
The MMN consists of three different-level memory modules, \ie, instance-level memory, part-level memory and domain-level memory, which are beneficial for discovering the similarity-and-dissimilarity relationships from fine-grained information to holistic representations in the target domain.
These three memory modules provide different yet complementary representations with each other while cooperatively improve the model generalization.
Experiments demonstrate that our MMN achieves competitive performance on three large datasets.
In the feature, we plan to extend our MMN to other domain adaptive tasks, such as face recognition and semantic segmentation.

\section{Appendix}
\subsection{More Experiments for Parameter Analysis}
\textbf{Analysis of Different Settings of $k$ for the Selected $k$-Nearest Neighbors.}
The hyper-parameter $k$ is used in Eq.~$\left(\textcolor{red}{2}\right)$ and Eq.~$\left(\textcolor{red}{8}\right)$ in the main paper, which defines the number of the most relevant samples for a given query image.
As demonstrated in Table~\ref{tab:k-nearest}, the performance improves with the increase of $k$ first and achieves the relatively best result when $k$ is set as $5$ to $15$.
With the increase of $k$ further, the performance decreases.
We believe that it is because the model with a larger $k$ could involve more false-positive samples, which hampers the model training.
In this paper, we set $k=10$ in all experiments except this part.

\begin{table}[h!]
\small
\setlength{\belowcaptionskip}{-0.1cm}
\setlength{\abovecaptionskip}{-0.4cm}
\begin{center}
\setlength{\tabcolsep}{2.0mm}{
\begin{tabular}{c|c|c|c|c}
\hline
\multirow{2}{*}{Number of the $k$} & \multicolumn{2}{c|}{D$\rightarrow$M} & \multicolumn{2}{c}{M$\rightarrow$D} \\ 
\cline{2-5} 
 & mAP & Rank-1 & \multicolumn{1}{c|}{mAP} & Rank-1 \\ 
\hline
\hline
1 & 63.7 & 85.1 & 52.1 & 72.1 \\
5 & 63.4 & 85.5 & 53.4 & 72.6   \\ 
10 & \textbf{65.1}  & \textbf{86.0} & \multicolumn{1}{c|}{53.9} & 73.2   \\
15 & 64.2 & 84.8 & {\textbf{54.5}} & \textbf{73.6}  \\ 
20 & 62.3 & 83.5 & 53.8 & 72.9   \\
25 & 61.0 & 83.4 & 53.4 & 72.9   \\
30 & 60.8 & 83.2 & 51.9 & 71.6   \\
\hline
\end{tabular}}
\end{center}
\caption{Evaluation with different number of the $k$-nearest neighbors on D$\rightarrow$M and M$\rightarrow$D. M denotes Market-1501~\cite{bai2017scalable} dataset while D denotes DukeMTMC-Re-ID~\cite{zheng2017unlabeled} dataset.}
\label{tab:k-nearest}
\end{table}

\begin{figure}[h!]
\centering
\includegraphics[trim =0mm 0mm 0mm 0mm, clip, width=.95\linewidth]{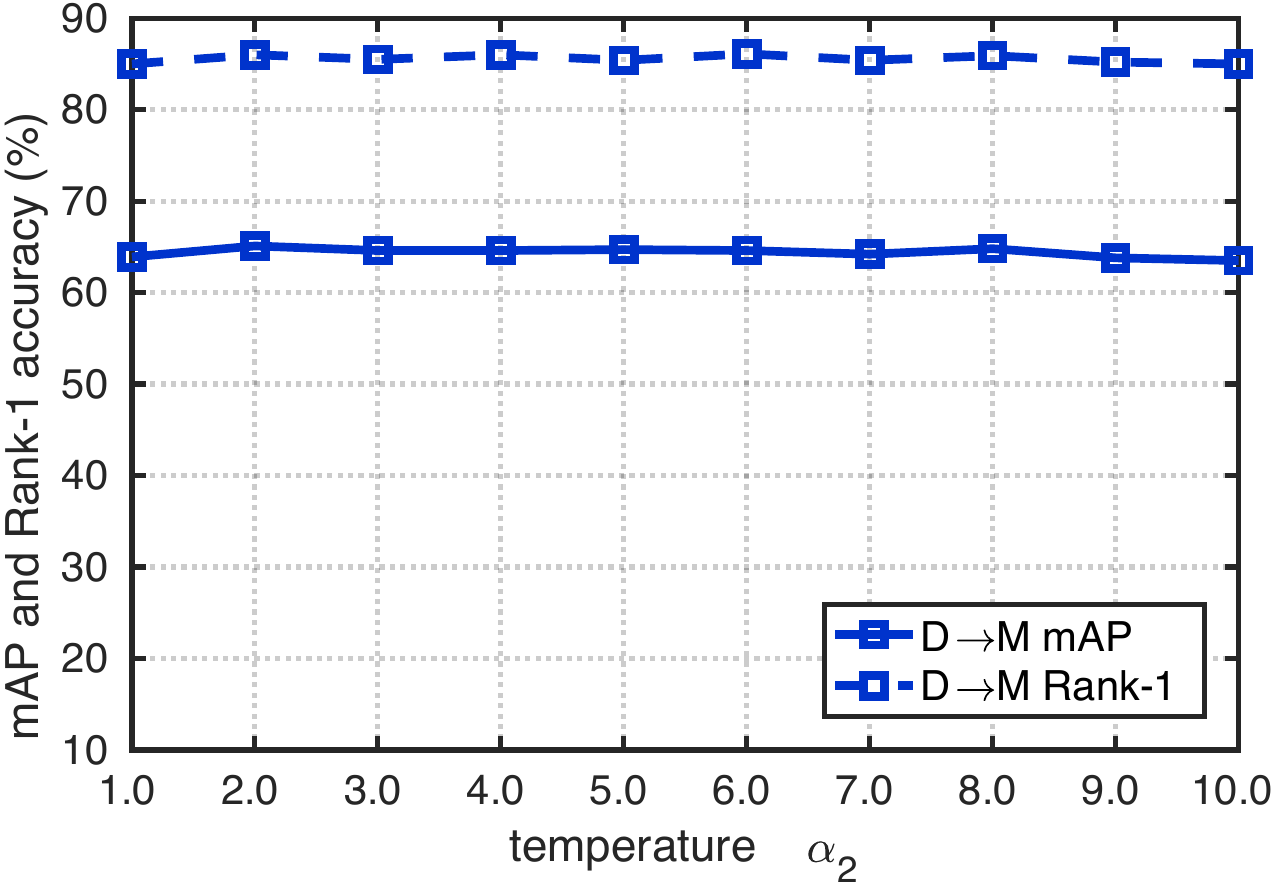}
\setlength{\abovecaptionskip}{0.2cm} 
\setlength{\belowcaptionskip}{-0.3cm}
\caption{
The influence of temperature factor $\alpha_2$ on D$\rightarrow$M.
}
\label{fig:temperature}
\end{figure}

\textbf{Analysis of Temperature Factor $\alpha_2$ in the Cosine Similarity Measurement.}
As described in Eq.~$\left(\textcolor{red}{7}\right)$, we use the temperature $\alpha_2$ to control the importance of the selected $k$ samples.
We conduct several experiments to evaluate the sensitivity of our method to $\alpha_2$ when transferring the model from DukeMTMC-Re-ID~\cite{zheng2017unlabeled} to Market-1501~\cite{bai2017scalable}.
From Figure~\ref{fig:temperature}, we can observe that when setting $\alpha_2$ in the range of $2.0$ to $8.0$, our method can obtain consistently high performance, and the best result is achieved when $\alpha_2=2.0$.
When $\alpha_2$ is too large, the mAP and Rank-$1$ accuracy will decrease simultaneously since the calculated score does not reflect the difference among images obviously enough.
When $\alpha_2=1$, the mAP is impacted marginally, and the Rank-1 accuracy is still high.
It shows that our method is insensitive to $\alpha_2$ when $\alpha_2$ is in an appropriate range.

\subsection{More Qualitative Analyses}
\begin{figure*}[tbp]
\centering
\includegraphics[trim =0mm 0mm 0mm 0mm, clip, width=1\linewidth]{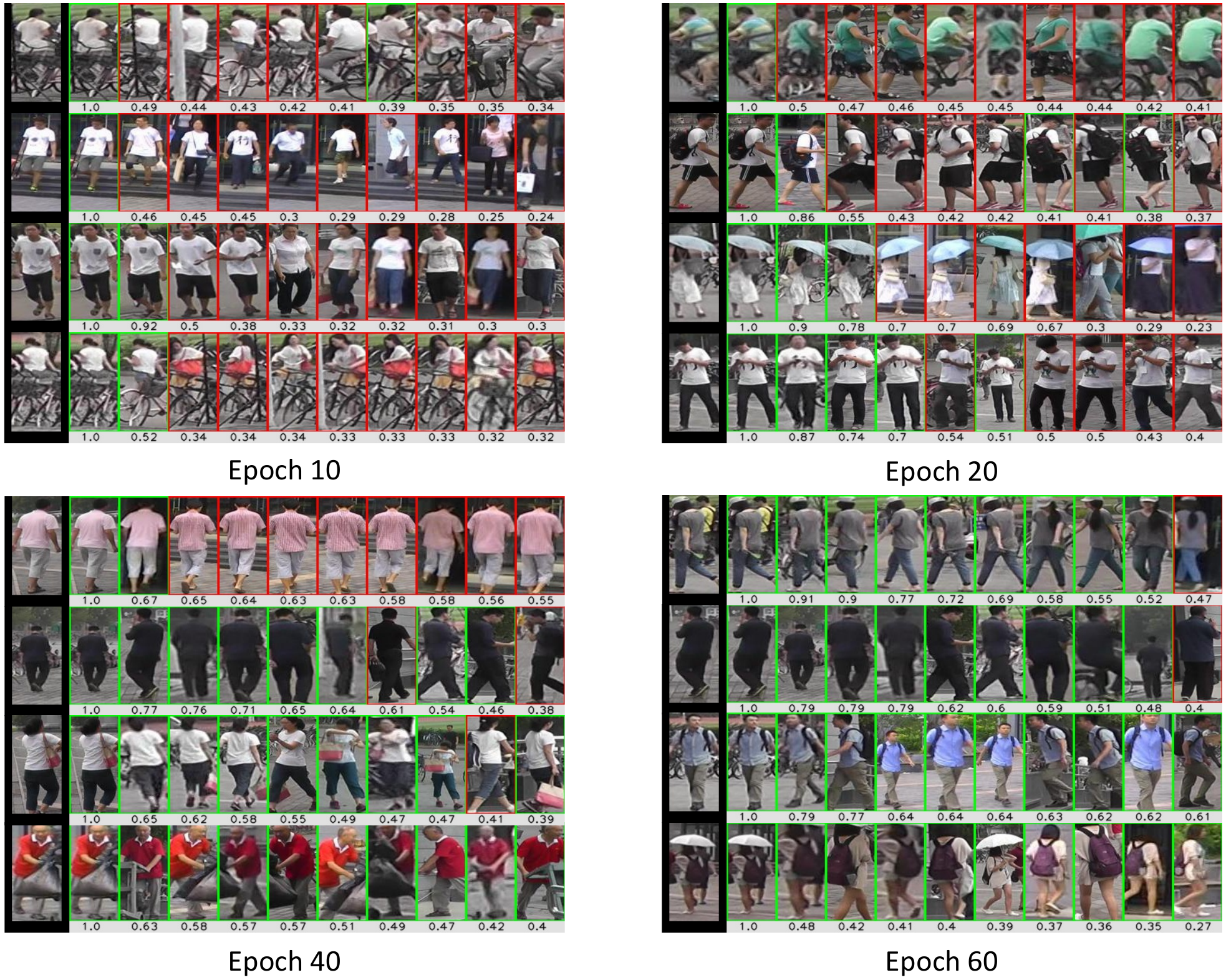}
\caption{Qualitative Analysis of the Selected $k$-Nearest Neighbors by the visualization on the Market-1501~\cite{bai2017scalable} target training data.
We choose the top-$10$ nearest neighbors for each query image and illustrate the images at epoch $10$, $20$, $40$ and $60$.
The first image in each row is the query image.
The \textcolor{green}{\textbf{green rectangle}} means images from the same identity with the query image, and the \textcolor{red}{\textbf{red rectangle}} represents the images from different identities.
The value at the bottom of each image is the similarity between the selected image and the query image.
It is clear that our MMN can improve the quality of the selected images along with the training process.}
\label{fig:multitargets}
\end{figure*}

\textbf{Qualitative Analysis of the Selected $k$-Nearest Neighbors.}
To demonstrate the results intuitively, we visualize the selected $k$-nearest neighbors along with the training process.
In Figure~\ref{fig:multitargets}, we illustrate four query images and their top-$10$ nearest neighbors in each epoch.
We can see that our MMN framework can progressively improve the quality of the selected nearest neighbors with the training process going on.
For instance, in the early epoch $10$, the model can make mistakes when the appearance or the background of images is similar. 
As shown in the second row at epoch $10$, the model even cannot distinguish the images with different color upper clothes.
When the training is in epoch $40$, it clearly shows that the difficulties of the model are mainly on subtle details, such as the bags and the strips of clothes.
When the training keeps going, we can further improve the ability of the model to focus on more detailed information and more robust on the problem of the misalignment.
It means that our MMN framework is beneficial for learning the similarity-and-dissimilarity of the training images with the fine-grained part information and the global structure of the whole training dataset.

\begin{figure*}[t!]
\centering
\includegraphics[trim =0mm 0mm 0mm 0mm, clip, width=.975\linewidth]{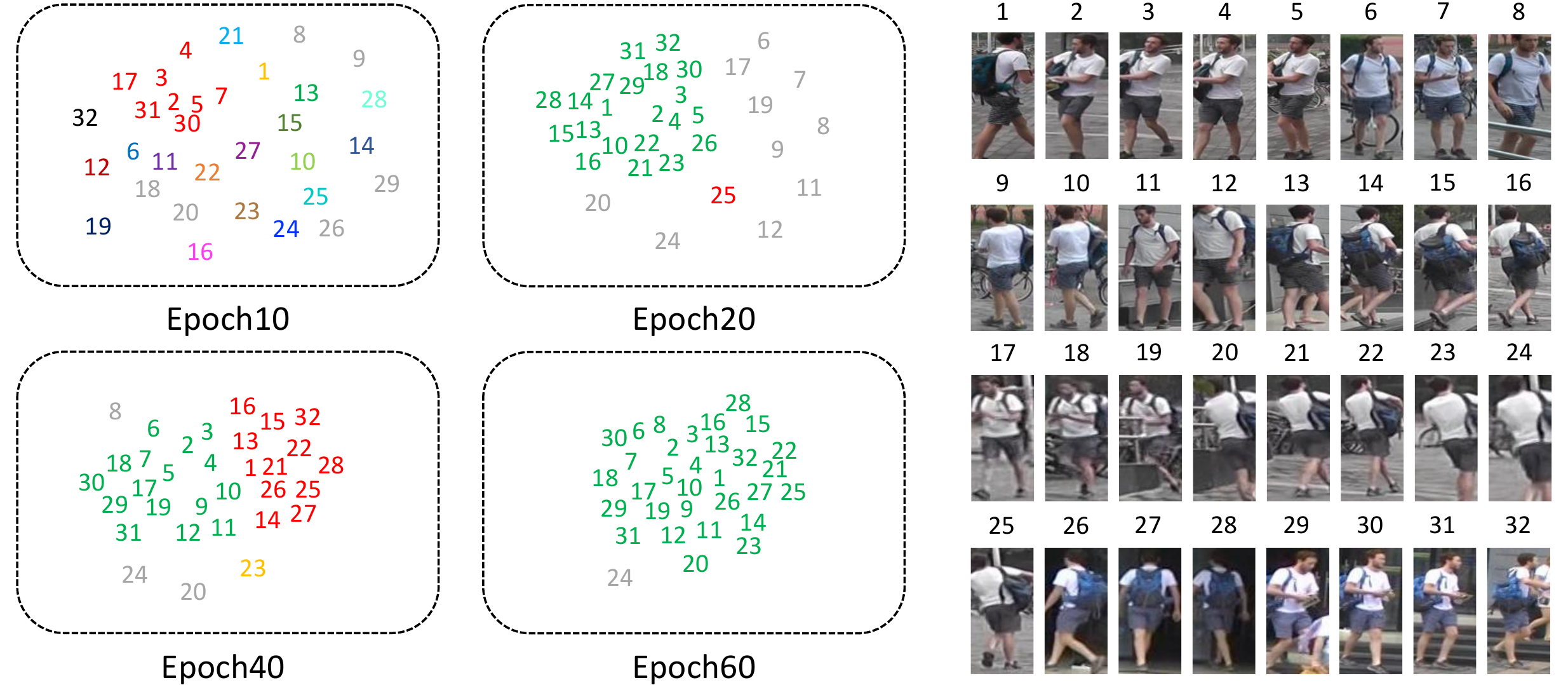}
\caption{Analysis of the Clustering Quality During the Training. All the $32$ images are from the same identity with no labels in this scenario, which has a large intra-class variance.
We illustrate the clustering results on epoch $10$, $20$, $40$ and $60$.
Images with same color belong to the same cluster.
\textcolor{gray}{Gray} images mean the samples do not belong to any cluster and are not used for training.
From epoch $10$ to epoch $60$, we can see that more images are used for training and the quality of the clustering result becomes more reliable with the training going on.}
\label{fig:clusters}
\end{figure*}

\textbf{Analysis of the Clustering Quality During the Training.}
We also analyze the clustering quality along with the training process.
As illustrated in Figure~\ref{fig:clusters}, the clustering quality becomes more reliable since our MMN improves the feature representation.
Moreover, we can observe that more samples are selected for training.
In the early stage, some images with different appearances are ignored by the clustering since the feature representations have huge variances compared with others for each real identity.
With the training going on, images with similar appearances are grouped, and more images are used for training.
It is because that the extracted feature representations are more discriminative.
Based on the consideration of the part information and the global structure simultaneously, our MMN can further improve the generalization of the model with more reliable features, and the images with large intra-variances can be better resolved.

{\small
\bibliographystyle{ieee_fullname}
\bibliography{egbib}
}

\clearpage

\end{document}